\documentclass[runningheads]{llncs}
\usepackage[T1]{fontenc}
\usepackage{float}
\usepackage{amssymb}
\usepackage{hyperref}
\usepackage{multirow}
\hypersetup{hypertex=true,
colorlinks=true,
linkcolor=blue,
urlcolor=blue,
anchorcolor=blue,
citecolor=blue}
\usepackage{lineno}
\usepackage{graphicx}

\begin{document}
\title{ChrSNet: Chromosome Straightening using Self-attention Guided Networks\thanks{These authors contributed equally to this work.}}
\titlerunning{ChrSNet: Chromosome Straightening Networks}

\author{Sunyi Zheng\inst{1,2,*} \and 
Jingxiong Li\inst{1,2,3,*}\and
Zhongyi Shui\inst{1,2}\and
Chenglu Zhu\inst{1,2}\and
Yunlong Zhang\inst{1,2,3}\and
Pingyi Chen\inst{1,2,3}\and
Lin Yang\inst{1,2}}

\authorrunning{S. Zheng et al.}
\institute{Artificial Intelligence and Biomedical Image Analysis Lab, School of Engineering, Westlake University \and
Institute of Advanced Technology, Westlake Institute for Advanced Study \and
College of Computer Science and Technology, Zhejiang University
\email{yanglin@westlake.edu.cn}}

\maketitle              

\begin{abstract}
Karyotyping is an important procedure to assess the possible existence of chromosomal abnormalities. However,  because of the non-rigid nature, chromosomes are usually heavily curved in microscopic images and such deformed shapes hinder the chromosome analysis for cytogeneticists. In this paper, we present a self-attention guided framework to erase the curvature of chromosomes. The proposed framework extracts spatial information and local textures to preserve banding patterns in a regression module. With complementary information from the bent chromosome, a refinement module is designed to further improve fine details. In addition, we propose two dedicated geometric constraints to maintain the length and restore the distortion of chromosomes. To train our framework, we create a synthetic dataset where curved chromosomes are generated from the real-world straight chromosomes by grid-deformation. Quantitative and qualitative experiments are conducted on synthetic and real-world data. Experimental results show that our proposed method can effectively straighten bent chromosomes while keeping banding details and length.

\end{abstract}

\section{Introduction}

Metaphase chromosome analysis is a fundamental step in the investigation of genetic diseases such as Turner syndrome, Down syndrome, and Cridu chat syndrome \cite{jorgensen2019comorbidity,theisen2010disorders}. Chromosomes tend to be curved in random degrees in the stained micro-photographs because of their non-rigid nature \cite{song2021novel}. Such a morphological deformation pose difficulties for chromosome analysis. Straightening of chromosomes could alleviate this problem and, therefore, has the potential to boost the efficiency of karyotyping for cytogeneticists. 

Prior works designed for chromosome straightening are mainly based on geometry. These geometric methods can be divided into two directions: medial axis extraction and bending point detection. For the first direction, Somasundaram et al. \cite{somasundaram2014straightening} employed the Stentiford thinning algorithm to select the medial axis and the projective straightening algorithm for axis correction. The axis was then considered as a reference to reconstruct straight chromosomes by projection. Similarly, Arora et al. \cite{arora2017algorithm} acquired the medial axis via morphological operations, and details were interpolated iteratively along the medial axis to produce straight chromosomes. Although the straightening methods using medial axis extraction can erase the curvature of bent chromosomes, some banding patterns can be distorted. Another type of method took the advantage of bending points to straighten chromosomes. For example, Roshtkhari et al. \cite{roshtkhari2008novel} and Sharma et al. \cite{sharma2017crowdsourcing} utilized bending points to separate a chromosome into parts. Rotation and stitching were applied to these parts for the reconstruction of straight chromosomes. However, these methods can cause discontinued banding patterns and inconsistent length after straightening. Besides, this method relied on bending points and inaccurately located points can severely harm the straightening performance.
 
Deep learning techniques have reached great achievements in the classification of chromosome types \cite{xiao2021deepacc,qin2019varifocal,zhang2018chromosome}, image translation \cite{eslami2020image,armanious2020medgan,armanious2019unsupervised} and portrait correction \cite{tan2021practical,zhu2021semi}, but rare studies have successfully applied deep learning in chromosome straightening. One of the challenges is that it is difficult to collect images of the same chromosome with both straight and curved shapes in the real-world to build a deep learning based mapping model for chromosome straightening. Recently, Song et al. \cite{song2021novel} developed a model utilizing conditional generative adversarial networks to erase curvatures in chromosomes. Their model created mappings between bent chromosomes and synthetic bent backbones. After model training, the model inversely transformed synthetic straight backbones to straightened chromosomes. Although banding patterns can be effectively restored by this method, the length of the straightened chromosome is inconsistent with the curved one's. Furthermore, the customized model was required to be implemented every time for each new chromosome. The feasibility of applying such a model in clinical practice needs to be further considered. To date, while many methods have been developed for chromosome straightening, challenges including limited model generalizability, lack of paired images for training, inconsistent length and inaccurate patterns after straightening remain in this task.

To this end, we propose a novel chromosome straightening approach using self-attention guided networks. The developed method combines low-level details and global contexts to recover banding patterns. A refinement module is applied to further recover fine details using the input bent chromosome as complementary information. Moreover, we design dedicated geometric constraints in terms of length and straightness to optimize the straightening process. In order to train our model, we propose a non-rigid transformation strategy that synthesizes curved chromosomes from the real-world straight ones. Note that we are the first study that creates mappings between straight and curved chromosomes for chromosome straightening. Our experiments show that the developed framework can effectively straighten various types of chromosomes with preserved banding details and length on both synthetic and real-world data.

\section{Methods}
The proposed approach for chromosome straightening mainly consists of two major steps. We first create a chromosome synthesizer that produces curved chromosomes from straight ones which exist in the real-world. With the use of curved and straight chromosome pairs, a framework is then trained for the removal of chromosome distortions. Methods are described in detail as follows:
\subsection{Curved chromosome synthesizer}
To build up a framework for chromosome straightening, our intuition is to create mappings between bent and straight chromosome pairs. However, acquiring such pairs from the same chromosome is challenging under a microscope. Moreover, chromosomes can be deformed randomly and it is nearly impossible to collect sufficient data that covers all possible situations in the real-world. 

Studies show that synthetic data generation could provide a solution for the lack of data in training deep learning methods \cite{ye2021multi,pan2021collaborative}. Inspired by this idea, we create a chromosome synthesizer that uses straight samples selected from a public chromosome dataset \cite{poletti2008automatic} to emulate their corresponding curved chromosomes. In the real-world, distortions of chromosomes can be a mixture of curvatures. To mimic real-world situations as closely as possible, two common types of distortions, slight curves and strong curves, are carefully defined on distortion meshes which are employed to convert straight chromosomes to bent ones. Detailed procedures for the generation of bent chromosome are composed as follows.

\noindent
\subsubsection{Deformation modeling. }When wrapping a chromosome image, a mesh $M$ with size $H \times W$ is given with a control point $p$ that is randomly selected from points on the skeleton of a straight chromosome. 
A random vector $v$, representing the deformation direction, is then generated to be perpendicular to the skeleton.The deformation function is defined as:
\begin{equation}
\label{eq1}
    p' = p + \omega v
\end{equation}
where $p'$ stands for the vertex after deformation. The factor $\omega$ controls the strength of deformation and it attenuates with the increasing distance $d$ between $p$ and the straight line defined by $p$ and $v$. The definition of $\omega$ for slight and strong curves could be expressed as:
\begin{equation}
\omega_{sli}=1-d^\alpha
\end{equation}
\begin{equation}
\omega_{str}=\frac{\alpha}{d+\alpha}
\end{equation}
where distance $d$ is normalized and constant $\alpha$ controls the propagation of deformation. A small $\alpha$ leads to a limited deformation only affecting in a local area around $p$, whereas a large $\alpha$ causes a global deformation. 

\subsection{ChrSNet: Chromosome straightening networks}
Chromosome straightening could be treated as a regression task that aims to provide pixel-to-pixel predictions from an image $x$ with size $H \times W \times C$ to a target $y\in \mathbf{R}^{H \times W \times C}$. This inspires us to follow the classic encoder-decoder architecture using a convolutional neural network (CNN). In spite of CNN's superiority on the extraction of local textures, it might not be able to recover spatial features in the straightening task due to CNN's drawback on modeling explicit long-range relationships \cite{chen2021transunet}. In contrast, by evaluating image through tokenized patches globally, transformer learns spatial information which can be used to reduce regression uncertainty. Hence, we build up a straightening model by importing self-attention via transformer layers which are implemented in a CNN-based encoder-decoder architecture. To produce straightened chromosomes with consistent morphological features and structure details, customized loss functions are also designed.

\subsubsection{Network architecture. }
The overview of our proposed framework ChrSNet is illustrated in Fig.1. The framework could be divided into a regression module and a refinement module. Specifically, in the regression module, a curved chromosome image $x$ is first down-sampled by a CNN encoder to create a feature map $f_M$ with size $h \times w \times c$. Then we divide $f_M$ into patches with the same size and flatten them into a one-dimensional sequence. These patches are denoted as $\left\{f_t^i \in \mathbf{R}^{p \times p \times c} \mid i=1,2,3...N_{k}\right\}$, where $p \times p$ is the size of one patch and $N_k=\frac{hw}{p^2}$ is the number of tokenized patches. Afterwards, patches are mapped into a latent embedding space via a linear projection. To encode the spatial information, patch embeddings and specific position embeddings are merged to generate the encoded image representation $z$ that is forwarded into multiple transformer layers. These embeddings are combined with low-level features by skip-connections to construct the preliminary mesh $M_1$. Since $M_1$ can be unsatisfactory, UNet \cite{ronneberger2015u} is employed as a refinement module to further improve the detailed information. In this module, we apply the input $x$ here to provide the complementary information. The input is concatenated with $M_1$ to generate $M_2$ which is refined again to acquire $M_3$. The final result $y$ can be acquired by applying $M_3$ on $x$.

\begin{figure}[tb]
\includegraphics[width=\textwidth]{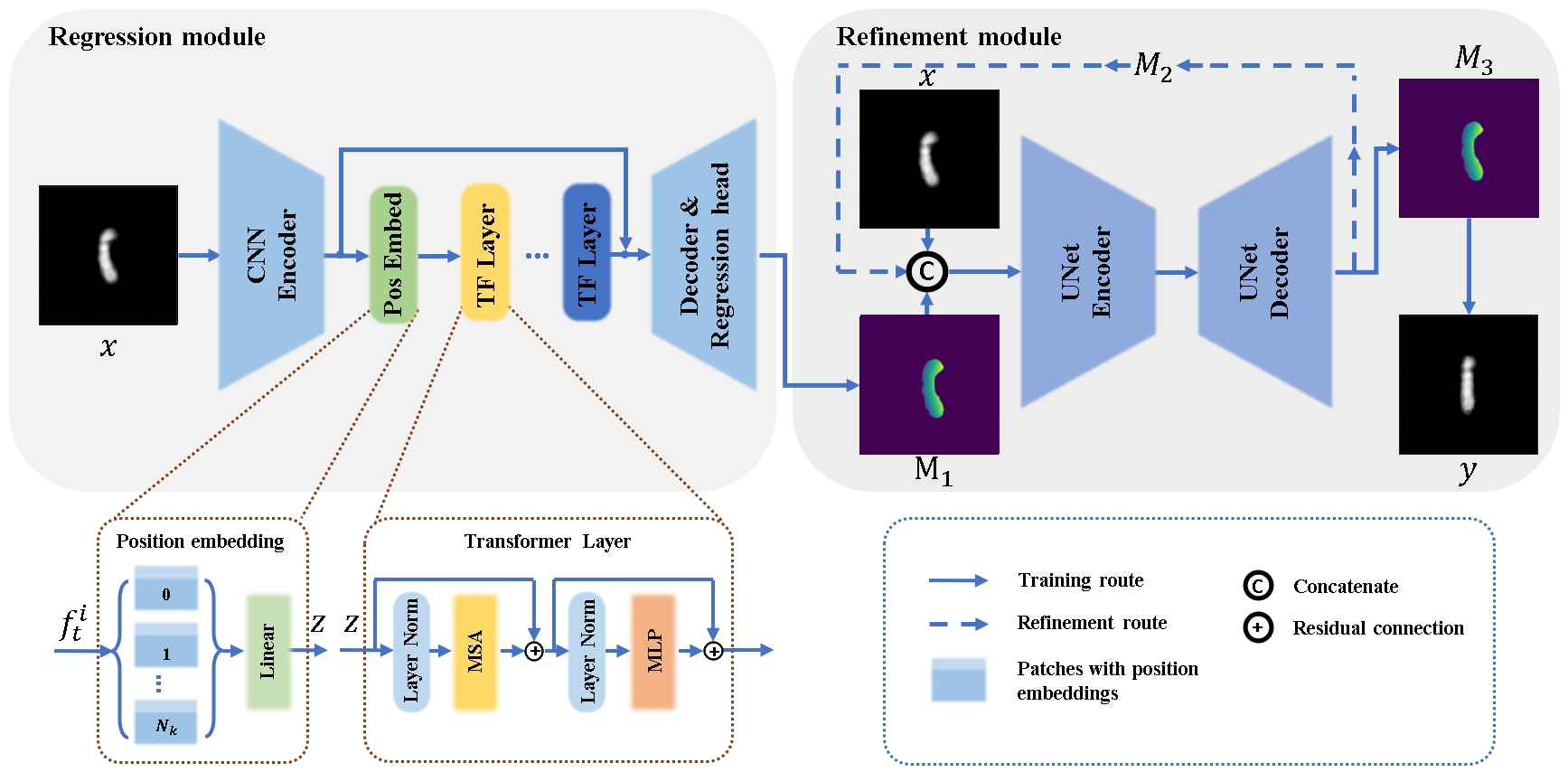} 
\label{fig1}
\caption{Overview of the proposed architecture. The architecture consists of a regression module using self-attention and a refinement module to improve fine details. A curved chromosome image $x$ is first fed into the regression module to obtain a rough mapping mesh $M_1$. Then the combination of $M_1$ and $x$ is fed into the refinement module to produce $M_2$ that is refined again for the generation of $M_3$. At last, the chromosome is straightened by applying $M_3$ on $x$.}
\end{figure}

\subsubsection{Loss function. }
A hybrid loss function is designed with constraints in the aspects of structure and texture to ensure the quality of the straightening process.

Regarding the structure loss, length and straightness are considered.  
In the real-world, chromosomes are normally curved rather than stretched because of their non-rigid nature. The straightened chromosome $y$ should have the same length as the deformed case $x$. To estimate the length of a chromosome, we count the number of pixels in the skeletonized chromosome $I_s$. As a result, the length preserving loss $L_{len}$ can be written as:

\begin{equation}
L_{len} = \frac{\left | l-l' \right|}{l'}
\end{equation}

\noindent
where $l$ is the predicted length and $l'$ is the target length. 

In addition to the consistent length after straightening, the skeleton of the chromosome should be straight. For this purpose, we propose a straightness loss $L_{str}$ for chromosome straightening. It is calculated by the overall slope (OS) and slope variations with uniformly sampled $n$ points $s_0, s_1, ..., s_n$. The straightness loss function can be presented as:
\begin{equation}
OS=\frac{y_{s_{0}}-y_{s_{n}}}{x_{s_{0}}-x_{s_{n}}}
\end{equation}
\begin{equation}
L_{str}= \frac{1}{n} \sum_{i=1, \ldots, n}\left[\frac{y_{s_{i}}-y_{s_{i-1}}}{x_{s_{i}}-x_{s_{i-1}}}-OS\right]
\end{equation}
where the coordinates of the point $s_i$ are ($x_{s_i}$, $y_{s_i}$) and the total number of sample points $n$ is set to $6$.

Preservation of banding textures is also important for later analysis including chromosome classification and abnormality identification. To keep texture features consistent after straightening, we use the $L1$ loss as the texture loss:
\begin{equation}
L_{tex}=\left | y-y' \right|
\end{equation}
The overall loss function is a combination of $L_{len}$, $L_{str}$ and $L_{tex}$ with the coefficient of $\alpha, \beta$ and $\theta$. The overall loss function is expressed as:
\begin{equation}
L_{all}=\alpha L_{tex} + \beta L_{len} + \theta L_{str}
\end{equation}

\section{Experiments}
\subsubsection{Evaluation metrics. }
We evaluate the model performance in terms of feature similarity and structure consistency. Due to morphological deformation between curved chromosomes and straightened results, commonly used metrics, such as Euclidean distance, structural similarity index \cite{wang2004image}, might not be appropriate for the evaluation of feature similarity. By contrast, Learned Perceptual Image Patch Similarity (LPIPS) \cite{zhang2018perceptual} is a neutral metric and measures a perceptual distance between two similar images using the way close to human judgments. Therefore, we employ LPIPS to evaluate whether the straightened chromosomes contain image features consistent with their corresponding curved chromosomes.

In the clinic, length can determine types of autosomes and curvature reflects deformation levels which affects difficulties for pattern analysis. Both of them are critical factors for later analysis, such as autosome classification. Hence, we assess the structure consistency by a length score (L score) and a straightness score (S score), where L score $= 1-L_{len}$, S score $ = 1-L_{str}$. An increasing L score means the length is more consistent after straightening, while a larger S score represents the corrected chromosome has a more straight shape. 

\noindent
\subsubsection{Dataset settings. }
Every human cell normally has 46 chromosomes and curvatures are often observed on chromosomes with long arms. Therefore, this study focuses on the straightening of chromosomes $1$-$12$. We select $1310$ human chromosome images ($704$ straight and $606$ curved individuals) from a public dataset \cite{poletti2008automatic} for experiments. To create curved and straight chromosome pairs for training the straightening model, our proposed data generator produces 100 synthetic curved chromosomes with 1-3 control points randomly selected on each straight chromosome. Slight curving and strong curving distortions are applied in a ratio of $1:1$ for the generation of synthetic chromosomes with different curvatures. Before training or test, chromosome images are centered and unified to the size of $256 * 256$ with zero-padding. 

\noindent
\subsubsection{Implementation details. }
Experiments are implemented based on PyTorch using a NVIDIA A100 GPU. The synthetic dataset is randomly separated into a training, validation and test set in a ratio of $10:1:1$ for modelling. The straightening model is trained with an initial learning rate of $5 \times 10^{-4}$ gradually decreased to $1 \times 10^{-5}$. The optimizer is Adam and the batch size is $24$. The training stops if the performance on the validation set does not improve for $20$ epochs. Our code is available at \url{https://github.com/lijx1996/ChrSNet}. 

\begin{figure}[tb]
\setlength{\abovecaptionskip}{0.cm}
\setlength{\belowcaptionskip}{-0.cm}
\centering
\includegraphics[width=\textwidth]{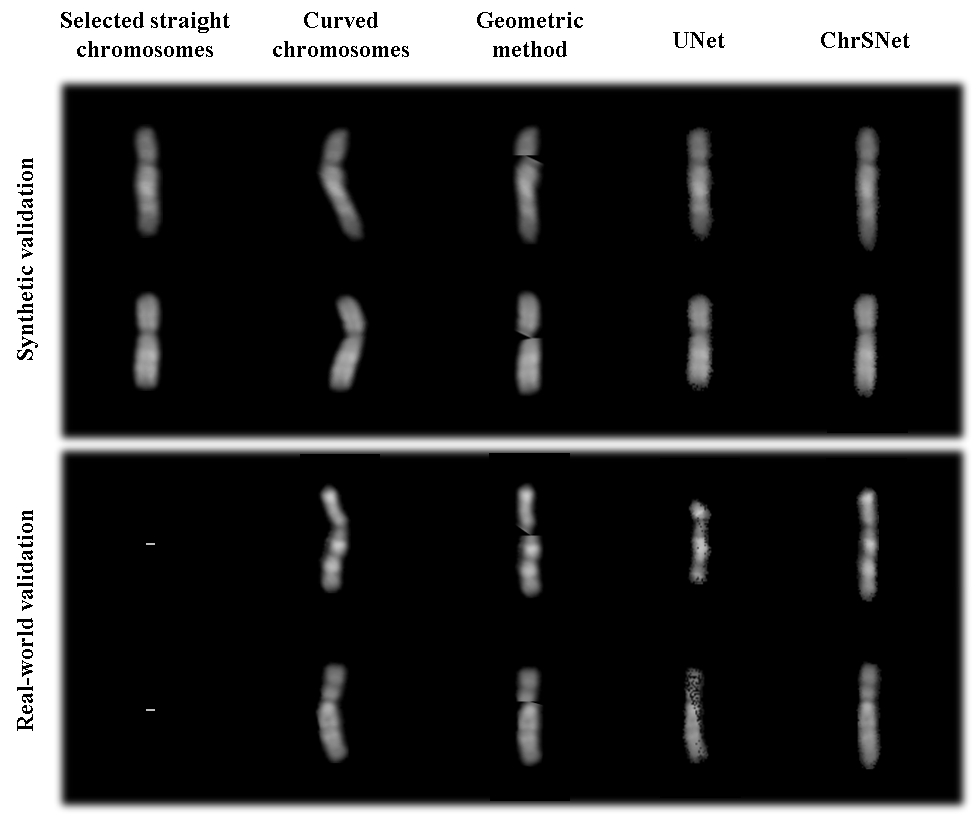} 
\caption{Examples of chromosomes straightened from the curved chromosomes by a geometric method \cite{roshtkhari2008novel}, UNet and ChrSNet on synthetic and real-world datasets. In the synthetic validation, straight chromosomes are selected from the real-world to generate curved chromosomes, whereas there is no corresponding straight chromosome for the curved chromosomes in the real-world validation.}
\label{fig2}
\end{figure}

\noindent
\subsubsection{Results and discussions. }
We perform comparisons among a geometric method \cite{roshtkhari2008novel}, UNet and our proposed method ChrSNet for chromosome straightening on synthetic and real-world data. Image examples of straightened chromosomes generated by each method on both sets are illustrated in Fig. \ref{fig2}. The results of different methods are presented in Table \ref{tab1:methods}. From the table, if straightness and length are considered, we can find that our designed method shows the best straightening performance on both datasets. This fact shows the superior performance of our method in the restoration of chromosome structures. When we only take the straightness into account, ChrSNet and UNet have competitive scores on both two datasets, but outperform the geometric method. In terms of length, although UNet performs much better than the geometry method on synthetic data, it fails to recover the length information on real data. Regarding the restoration of banding details, UNet has the highest LPIPS score of 89.29, similar to the score of ChrSNet on synthetic data, but its performance degrades in the real-world validation. The reason for the performance difference of UNet on two datasets could be that real-world data is more complex than synthetic data and generalizability of Unet is poor without self-attention on complex data. Conversely, ChrSNet achieves a much higher LPIPS value of 87.18 compared to UNet in real cases, which shows the model generalizability and effectiveness of ChrSNet in preserving banding patterns.

\begin{table}[tb]
\centering
\setlength{\abovecaptionskip}{0.cm}
\setlength{\belowcaptionskip}{0.cm}
\caption{Quantitative comparisons between our proposed ChrSNet and other methods on synthetic and real-world data. We assess the chromosome structure after straightening from straightness (S score) and length (L score). The learned perceptual image patch similarity (LPIPS) is applied to evaluate preserved banding details.}
\label{tab1:methods}
\begin{tabular}{c|c|c|c|c|c|c}
\hline
\multirow{2}{*}{\textbf{Models}}      & \multicolumn{3}{c|}{\textbf{Synthetic validation}} & \multicolumn{3}{c}{\textbf{Real-world validation}} \\ \cline{2-7} 
 &
  \multicolumn{1}{c|}{\begin{tabular}[c]{@{}c@{}}S score\\ \end{tabular}} &
  \multicolumn{1}{c|}{\begin{tabular}[c]{@{}c@{}}L score\\ \end{tabular}} &
  \begin{tabular}[c]{@{}c@{}}LPIPS\\ \end{tabular} &
  \multicolumn{1}{c|}{\begin{tabular}[c]{@{}c@{}}S sore\\ \end{tabular}} &
  \multicolumn{1}{c|}{\begin{tabular}[c]{@{}c@{}}L score\\ \end{tabular}} &
  \begin{tabular}[c]{@{}c@{}}LPIPS\\ \end{tabular} \\ \hline
\multicolumn{1}{l|}{Geometry} & 76.94$\pm$20.4    & 82.32$\pm$18.8   & 85.20$\pm$5.33   & 81.07$\pm$10.4   & 86.25$\pm$12.3   & 85.38$\pm$6.48  \\
UNet & 86.01$\pm$10.3    & 92.70$\pm$5.36   & \underline{89.29}$\pm$4.00   & 88.46$\pm$6.81   & 74.90$\pm$11.4   & 80.56$\pm$5.50  \\
ChrSNet                               & \underline{87.68}$\pm$7.95    & \underline{96.22}$\pm$4.49   & 88.75$\pm$4.17   & \underline{88.56}$\pm$5.98   & \underline{94.87}$\pm$5.05   & \underline{87.18}$\pm$5.36  \\ \hline
\end{tabular}
\end{table}

\begin{table}[tb]
\centering
\caption{Ablation study of our proposed network on two datasets. "RG", "RT", "SL", "RR" refer to the regression module, refinement module in the training route, structure loss and refinement process in the refinement route.}
\label{tab2:ablation}
\begin{tabular}{cccc|c|c|c|c|c|c}
\hline
\multicolumn{4}{c|}{\textbf{Components}}& \multicolumn{3}{c|}{\textbf{Synthetic validation}} & \multicolumn{3}{c}{\textbf{Real-world validation}} \\ \hline
\multicolumn{1}{c|}{RG} &
  \multicolumn{1}{c|}{RT} &
  \multicolumn{1}{c|}{SL} &
  RR &
  \multicolumn{1}{c|}{\begin{tabular}[c]{@{}c@{}}S score\\ \end{tabular}} &
  \multicolumn{1}{c|}{\begin{tabular}[c]{@{}c@{}}L score\\ \end{tabular}} &
  \begin{tabular}[c]{@{}c@{}}LPIPS\\ \end{tabular} &
  \multicolumn{1}{c|}{\begin{tabular}[c]{@{}c@{}}S sore\\ \end{tabular}} &
  \multicolumn{1}{c|}{\begin{tabular}[c]{@{}c@{}}L score\\ \end{tabular}} &
  \begin{tabular}[c]{@{}c@{}}LPIPS\\ \end{tabular} \\ \hline
\checkmark&       &       &      & 85.90$\pm$9.30 & 93.90$\pm$5.81 & \underline{90.01}$\pm$4.07 & 85.59$\pm$6.90& 93.13$\pm$5.38&\underline{89.02}$\pm$5.96\\
\checkmark&\checkmark&       &      & 85.78$\pm$8.89 & 94.22$\pm$5.19 & 89.70$\pm$4.05 & 87.05$\pm$6.50& 93.62$\pm$6.41&88.43$\pm$5.34\\
\checkmark&\checkmark&\checkmark&      & 86.14$\pm$9.32 & 95.35$\pm$4.58 & 89.10$\pm$4.01 & 87.44$\pm$6.57& 94.66$\pm$5.59&87.50$\pm$5.33\\
\checkmark&\checkmark&\checkmark&\checkmark   & \underline{87.68}$\pm$7.95 & \underline{96.22}$\pm$4.49 & 88.75$\pm$4.17 & \underline{88.56}$\pm$5.98& \underline{94.87}$\pm$5.05&87.18$\pm$5.36\\ \hline

\end{tabular}
\end{table}

We also assess the effect of four components in ChrSNet on both synthetic and real-world datasets (Table \ref{tab2:ablation}). First, baseline results are produced using the regression module with the texture loss only. Based on this, we verified the functionality of the refinement module in the training route, structure loss and refinement process in the refinement route. From Table \ref{tab2:ablation}, we can observe that S and L scores always show an increasing trend on both two sets after adding extra components. Based on a relatively high baseline, the S score improves by 2.3 at most, while the L score increases by 3.0 at most. Of note, standard deviations are smaller with more added components, which suggests the results become more stable. Therefore, adding extra architectures is helpful to improve model performance and robustness. Besides, the use of the regression module alone results in a LPIPS score of 90.01 (RG in Table 2) that is higher than the score of UNet (89.29 in Table 1) on synthetic data. In addition, as shown in Fig. \ref{fig2}, UNet without self-attention fails to recover correlated patterns at the bottom and top on real-world data, while ChrSNet using self-attention restores long-range patterns. These findings show that the use of self-attention is effective to keep pattern details in chromosome straightening. Although the LPIPS score slightly decreases when all components are combined, it is still comparable to the baseline result and therefore, will not significantly affect later analysis. 

\section{Conclusion}
In this study, we proposed a novel framework that utilized attention mechanism and convolutional neural networks to obtain global textures and local details for chromosome straightening. We also attempted to solve the limitations regarding length consistency and generalizability of the existing framework. Our method generated deformed chromosomes using two basic curving distortions and created direct mappings between straight and deformed chromosomes for straightening. The experimental results on synthetic and real-world data showed the effectiveness and robustness of our proposed approach in preserving banding details and structures on various types of chromosomes.

\bibliographystyle{splncs04}
\bibliography{paper1588}

\begin{thebibliography}{10}
\providecommand{\url}[1]{\texttt{#1}}
\providecommand{\urlprefix}{URL }
\providecommand{\doi}[1]{https://doi.org/#1}

\bibitem{armanious2019unsupervised}
Armanious, K., Jiang, C., Abdulatif, S., K{\"u}stner, T., Gatidis, S., Yang,
  B.: Unsupervised medical image translation using cycle-medgan. In: 2019 27th
  European Signal Processing Conference (EUSIPCO). pp.~1--5. IEEE (2019)

\bibitem{armanious2020medgan}
Armanious, K., Jiang, C., Fischer, M., K{\"u}stner, T., Hepp, T., Nikolaou, K.,
  Gatidis, S., Yang, B.: Medgan: Medical image translation using gans.
  Computerized medical imaging and graphics  \textbf{79},  101684 (2020)

\bibitem{arora2017algorithm}
Arora, T., Dhir, R., Mahajan, M.: An algorithm to straighten the bent human
  chromosomes. In: 2017 Fourth International Conference on Image Information
  Processing (ICIIP). pp.~1--6. IEEE (2017)

\bibitem{chen2021transunet}
Chen, J., Lu, Y., Yu, Q., Luo, X., Adeli, E., Wang, Y., Lu, L., Yuille, A.L.,
  Zhou, Y.: Transunet: Transformers make strong encoders for medical image
  segmentation. arXiv preprint arXiv:2102.04306  (2021)

\bibitem{eslami2020image}
Eslami, M., Tabarestani, S., Albarqouni, S., Adeli, E., Navab, N., Adjouadi,
  M.: Image-to-images translation for multi-task organ segmentation and bone
  suppression in chest x-ray radiography. IEEE transactions on medical imaging
  \textbf{39}(7),  2553--2565 (2020)

\bibitem{jorgensen2019comorbidity}
J{\o}rgensen, I.F., Russo, F., Jensen, A.B., Westergaard, D., Lademann, M., Hu,
  J.X., Brunak, S., Belling, K.: Comorbidity landscape of the danish patient
  population affected by chromosome abnormalities. Genetics in Medicine
  \textbf{21}(11),  2485--2495 (2019)

\bibitem{pan2021collaborative}
Pan, Y., Chen, Y., Shen, D., Xia, Y.: Collaborative image synthesis and disease
  diagnosis for classification of neurodegenerative disorders with incomplete
  multi-modal neuroimages. In: International Conference on Medical Image
  Computing and Computer-Assisted Intervention. pp. 480--489. Springer (2021)

\bibitem{poletti2008automatic}
Poletti, E., Grisan, E., Ruggeri, A.: Automatic classification of chromosomes
  in q-band images. In: 2008 30th Annual international conference of the IEEE
  engineering in medicine and biology society. pp. 1911--1914. IEEE (2008)

\bibitem{qin2019varifocal}
Qin, Y., Wen, J., Zheng, H., Huang, X., Yang, J., Song, N., Zhu, Y.M., Wu, L.,
  Yang, G.Z.: Varifocal-net: A chromosome classification approach using deep
  convolutional networks. IEEE transactions on medical imaging
  \textbf{38}(11),  2569--2581 (2019)

\bibitem{ronneberger2015u}
Ronneberger, O., Fischer, P., Brox, T.: U-net: Convolutional networks for
  biomedical image segmentation. In: International Conference on Medical image
  computing and computer-assisted intervention. pp. 234--241. Springer (2015)

\bibitem{roshtkhari2008novel}
Roshtkhari, M.J., Setarehdan, S.K.: A novel algorithm for straightening highly
  curved images of human chromosome. Pattern recognition letters
  \textbf{29}(9),  1208--1217 (2008)

\bibitem{sharma2017crowdsourcing}
Sharma, M., Saha, O., Sriraman, A., Hebbalaguppe, R., Vig, L., Karande, S.:
  Crowdsourcing for chromosome segmentation and deep classification. In:
  Proceedings of the IEEE conference on computer vision and pattern recognition
  workshops. pp. 34--41 (2017)

\bibitem{somasundaram2014straightening}
Somasundaram, D., Kumar, V.V.: Straightening of highly curved human chromosome
  for cytogenetic analysis. Measurement  \textbf{47},  880--892 (2014)

\bibitem{song2021novel}
Song, S., Huang, D., Hu, Y., Yang, C., Meng, J., Ma, F., Coenen, F., Zhang, J.,
  Su, J.: A novel application of image-to-image translation: Chromosome
  straightening framework by learning from a single image. In: 2021 14th
  International Congress on Image and Signal Processing, BioMedical Engineering
  and Informatics (CISP-BMEI). pp.~1--9. IEEE (2021)

\bibitem{tan2021practical}
Tan, J., Zhao, S., Xiong, P., Liu, J., Fan, H., Liu, S.: Practical wide-angle
  portraits correction with deep structured models. In: Proceedings of the
  IEEE/CVF Conference on Computer Vision and Pattern Recognition. pp.
  3498--3506 (2021)

\bibitem{theisen2010disorders}
Theisen, A., Shaffer, L.G.: Disorders caused by chromosome abnormalities. The
  application of clinical genetics  \textbf{3}, ~159 (2010)

\bibitem{wang2004image}
Wang, Z., Bovik, A.C., Sheikh, H.R., Simoncelli, E.P.: Image quality
  assessment: from error visibility to structural similarity. IEEE transactions
  on image processing  \textbf{13}(4),  600--612 (2004)

\bibitem{xiao2021deepacc}
Xiao, L., Luo, C.: Deepacc: Automate chromosome classification based on
  metaphase images using deep learning framework fused with priori knowledge.
  In: 2021 IEEE 18th International Symposium on Biomedical Imaging (ISBI). pp.
  607--610. IEEE (2021)

\bibitem{ye2021multi}
Ye, J., Xue, Y., Liu, P., Zaino, R., Cheng, K.C., Huang, X.: A multi-attribute
  controllable generative model for histopathology image synthesis. In:
  International Conference on Medical Image Computing and Computer-Assisted
  Intervention. pp. 613--623. Springer (2021)

\bibitem{zhang2018perceptual}
Zhang, R., Isola, P., Efros, A.A., Shechtman, E., Wang, O.: The unreasonable
  effectiveness of deep features as a perceptual metric. In: CVPR (2018)

\bibitem{zhang2018chromosome}
Zhang, W., Song, S., Bai, T., Zhao, Y., Ma, F., Su, J., Yu, L.: Chromosome
  classification with convolutional neural network based deep learning. In:
  2018 11th International Congress on Image and Signal Processing, BioMedical
  Engineering and Informatics (CISP-BMEI). pp.~1--5. IEEE (2018)

\bibitem{zhu2021semi}
Zhu, F., Zhao, S., Wang, P., Wang, H., Yan, H., Liu, S.: Semi-supervised
  wide-angle portraits correction by multi-scale transformer. arXiv preprint
  arXiv:2109.08024  (2021)

\end{thebibliography}

\end{document}